%% file: root.tex
\documentclass[letterpaper, 10 pt, conference]{styles/ieeeconf} 
\IEEEoverridecommandlockouts    
\usepackage{algorithm}
\usepackage{algorithmic}
\usepackage{subcaption}
\usepackage[dvipsnames]{xcolor}
\usepackage{graphicx}
\usepackage[bookmarks=false,colorlinks=true,pagebackref]{hyperref}

\input{include/packages.tex}
\newcommand{\etal}{\emph{et al.}}

\title{\textbf{
MVRackLay: Monocular Multi-View Layout Estimation for Warehouse Racks and Shelves
}}

\makeatletter
\renewcommand\AB@affilsepx{, \protect\Affilfont}
\makeatother

\author[1]{Pranjali Pathre\thanks{Corresponding author: \url{pranjali2000pathre@gmail.com} }}
\author[1]{Anurag Sahu}
\author[1]{Ashwin Rao}
\author[1]{Avinash Prabhu}
\author[1]{Meher Shashwat Nigam}
\author[1]{Tanvi Karandikar}
\author[2]{\\Harit Pandya}
\author[1]{K. Madhava Krishna} 

\affil[1]{\small{Robotics Research Center, IIIT Hyderabad, India}}
\affil[2]{\small{Toshiba Research, UK}}

\begin{document}
\bstctlcite{Force_Etal}

\makeatletter
\let\@oldmaketitle\@maketitle
\renewcommand{\@maketitle}{\@oldmaketitle
\centering

\includegraphics[width=\textwidth]{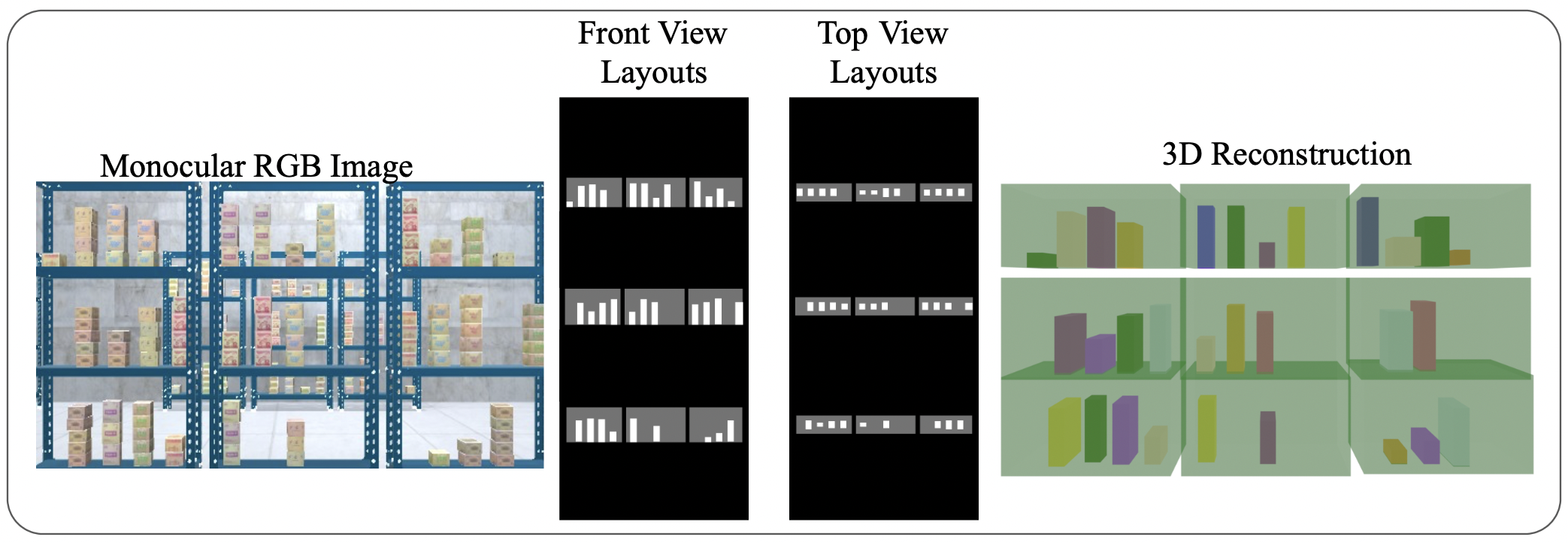}

\captionof{figure}{\small{
We introduce \textbf{MVRackLay}, a deep neural architecture, which, given a sequence of monocular RGB images of racks in a warehouse scene, predicts the \textit{top-view} \underline{and} \textit{front-view} occupancy layouts for all the racks and shelves, partly or completely visible, in a \textit{shelf-centric} frame.
Fusing these \textit{shelf-centric} layouts provides a 3D reconstruction of the rack, and stitching together all the racks across frames produces a 3D reconstruction of the entire warehouse.}
}
\label{fig:teaser}
\vspace{-3mm}
}

\makeatother

\maketitle

\begin{abstract}
In this paper, we propose and showcase, for the first time, monocular multi-view layout estimation for warehouse racks and shelves. Unlike typical layout estimation methods, MVRackLay estimates multi-layered layouts, wherein each layer corresponds to the layout of a shelf within a rack. Given a sequence of images of a warehouse scene, a dual-headed Convolutional-LSTM architecture outputs segmented racks, the front and the top view layout of each shelf within a rack. With minimal effort, such an output is transformed into a 3D rendering of all racks, shelves and objects on the shelves, giving an accurate 3D depiction of the entire warehouse scene in terms of racks, shelves and the number of objects on each shelf. MVRackLay generalizes to a diverse set of warehouse scenes with varying number of objects on each shelf, number of shelves and in the presence of other such racks in the background. Further, MVRackLay shows superior performance vis-a-vis its single view counterpart, RackLay \cite{racklay} in layout accuracy, quantized in terms of the mean IoU and mAP metrics. We also showcase a multi-view stitching of the 3D layouts resulting in a representation of the warehouse scene with respect to a global reference frame akin to a rendering of the scene from a SLAM pipeline. To the best of our knowledge, this is the first such work to portray a 3D rendering of a warehouse scene in terms of its semantic components - Racks, Shelves and Objects - all from a single monocular camera.

\end{abstract}
\setcounter{figure}{1} 
\input{sections/introduction}
\input{sections/relatedwork}
\input{sections/method}
\input{sections/dataset}
\input{sections/experiments}

\input{sections/conclusion}
\bibliography{references/IEEEexample}
\bibliographystyle{styles/IEEEtran}
\end{document}

%% file: include/packages.tex
\usepackage{times}
\usepackage{blindtext, rotating}

\usepackage{epsfig}
\usepackage{graphicx}
\usepackage{amsmath}
\usepackage{amssymb}

\usepackage{tabularx}

\usepackage[utf8]{inputenc}
\usepackage[T1]{fontenc}
\usepackage{textcomp}
\usepackage{gensymb}
\usepackage{multirow}
\usepackage{booktabs}
\usepackage{xcolor}
\usepackage{graphicx}
\usepackage{caption}
\usepackage{subcaption}
\usepackage[inline]{asymptote}
\usepackage{tikz}
\usepackage{subcaption}
\usepackage{float}
\usepackage{capt-of}
\usepackage{etoolbox}
\usepackage{adjustbox}
\usepackage[font=small]{caption}
\usepackage{paralist}

\usepackage[colorlinks=true, pagebackref]{hyperref}

\usepackage{epigraph}
\usepackage{authblk}
\DeclareMathOperator{\E}{\mathbb{E}}


%% file: sections/introduction.tex
\section{INTRODUCTION}
\label{sec:introduction}
The need for warehouse automation grows by the day, and in the future, a fleet of robots could manage an entire warehouse with little to no human intervention \cite{peter_2018}. Yet, almost 30\% of warehouses operate without their staple warehouse management systems  \cite{wmstrends}.
\looseness=-1

In this paper, we address the thus far untackled problem of \textit{multi-view\textit} layout estimation for all the visible racks in the image.
We propose a straightforward and effective network architecture \textit{MVRackLay}\footnote{Project page: \url{https://github.com/pranjali-pathre/vRacklay}}, which outputs the \textit{top-view} \underline{and} \textit{front-view} layouts of all shelves making up each rack, partly or wholly visible in every frame of an input sequence of monocular RGB images of a warehouse (these could be the frames of a video) (Fig. \ref{fig:teaser}). Note that a rack may only be partially visible in a frame. 
The network learns layouts in the canonical frame centered on the shelf, called the \textit{shelf-centric} layout.


An essential point to note is that the problem is not a direct application of a standard formulation of object recognition, semantic segmentation or layout estimation. While the above methods can be applied to objects on rack shelves \cite{unet, he2017mask}, present methods cannot be adapted directly to localize rack shelves, as shown in our baseline comparisons in Sec. \ref{sec:experiments:baselines}. While typical layout formulations estimate layouts with reference to a single dominant plane (such as the ground plane) \cite{roddick2018orthographic}, warehouse rack shelves take the form of disconnected planar segments, each present at different heights above the ground plane. Furthermore, they often appear occluded and diffused in warehouse scenes.
Thus, an important novelty of our formulation is the adaptation of deep architectures to the problem of multi-view layout estimation over multiple shelves and racks, as well as shelf contents. 

\textit{MVRackLay} leverages, using Convolutional LSTM layers, the temporal data across images and extends the layout prediction problem to a wider scope in a warehouse setting. Unlike \textit{RackLay\cite{racklay}}, where the model predicts the layouts only for the rack in focus, we design our model o predict the layouts for all the racks in the image, whether visible fully or partially in frame. Additionally, to leverage the temporal information across image sequences, we propose the multiview and multilayer layout prediction mechanism for \textit{MVRackLay}, where layouts are predicted over a sequence of images. Furthermore, through the downstream application of layout stitching across multiple views, we demonstrate that an explicit 3D model of the warehouse can be reconstructed from the predicted layouts. 

Real-world public datasets for warehouses are few and not comprehensive. To alleviate the issue of obtaining sufficient training data, we develop and open-source a complete synthetic data generation pipeline \textit{WareSynth}, an extended and improved version of the pipeline introduced in \cite{racklay}. Using domain randomization, \textit{WareSynth} can generate a huge variety of synthetic warehouses, suitably emulating any real-world warehouse. The warehouse dataset generation pipeline allows users to customize scenes as per their needs, automating both the image-capturing process as well as the generation of ground-truth annotations for these images. We train and evaluate the performance of \textit{MVRackLay} on such synthetic warehouse scenes. The monocular image sequences obtained from \textit{WareSynth} involve translation of the camera along a predefined path, facing a row of racks and at a fixed distance from them. The annotations for these sequences are fed into the deep \textit{MVRackLay} network.

\noindent Specifically, the paper contributes as follows: 
    \\ \textbf{1)} It presents a notable improvement over the formerly proposed \textit{RackLay} \cite{racklay} architecture (Sec. \ref{sec:method:arch}), the keynote of which is the use of Convolutional LSTM Layers, which enable the network to train on a monocular image sequence, rather than discrete images of racks. The network uses this previously-missing spatial data to improve \textit{shelf-centric} layout (Sec. \ref{sec:method:problemform}) predictions in each frame, specifically when racks and shelves are only partially visible in the image. \looseness=-1
    
    \noindent \textbf{2)}  We open-source a flexible data generation pipeline \textit{WareSynth} with domain randomization capabilities that empower a huge variety of synthetic data. We also release relevant instructions that enable the user to create and customize their warehouse scenes and generate 2D/3D ground truth annotations needed for their task automatically, as discussed in Sec. \ref{sec:dataset}. 
   
    \looseness=-1
   
    \noindent \textbf{3)} Further, we show several applications using these layout representations. Layout-enabled multi-view 3D warehouse reconstruction is a novel application discussed in Sec. \ref{sec:experiments:applications}. The downstream task of free space estimation can also be achieved. Most importantly, we show that the use of Convolutional LSTM layers to predict 3D representations of racks in each frame enables them to be stitched together into a 3D reconstruction of the warehouse in a global reference frame, similar to the rendering of a scene from a SLAM pipeline.
    
    \noindent \textbf{4)} We demonstrate significant performance gain on popular rubrics compared to previous methods \cite{shi2019pointrcnn, he2017mask} adapted to the estimation of shelf layouts. Equally important, we showcase notable improvement in layout estimation over the single-view estimator \cite{racklay}. Moreover, we compile and present results of several ablations involving variations in architecture which establish the superiority of \textit{MVRackLay} (Sec. \ref{sec:experiments}). 

%% file: sections/relatedwork.tex
\section{RELATED WORK}
\label{sec:relatedwork}

\noindent \textbf{Object detection methods:}
A significant portion of our problem deals with localizing semantic classes like shelves and boxes/cartons in a 3D scene. There exist several approaches to detect object layouts in 3D. Some of these \cite{ku2018joint, liang2018deep} combine information from images and LiDAR, while others \cite{roddick2018orthographic, wang2019pseudo} first convert images to \textit{bird’s eye view} representations, followed by object detection.

\noindent \textbf{Bird's eye view (BEV) representation:}
Schulter \etal \cite{schulter2018learning} proposed one of the first approaches to estimate an occlusion-reasoned BEV road layout from a single color image. Wang \etal \cite{wang2019parametric} build on top of \cite{schulter2018learning} to infer parameterized road layouts.  In contrast, our approach is non-parametric and hence, more flexible than such parametric models, which may not account for all possible layouts.
We take inspiration from MonoLayout \cite{monolayout} (which can be trained end to end on color images, reasons beyond occlusion boundaries and being non-parameterized, need not be actuated with these additional inputs) and extend it to multiple planes. 

\noindent \textbf{Single-view layout estimation:} 
\textit{RackLay} \cite{racklay} proposed a layout estimation technique that is able to predict shelf layouts of one rack at a time, which must be in focus and completely visible in a single monocular image. Our network \textit{MVRackLay} is more flexible in that it predicts shelf layouts of all fully-visible and partially-visible racks in a monocular image sequence and more accurate due to the incorporation of spatial data of warehouse racks from the consecutive frames of the input video.

\noindent \textbf{Warehouse Datasets:} 
\label{sec:relatedwork:dataset}
Publicly available datasets for warehouse settings are far and in between. Real-world datasets like LOCO \cite{loco2020} exist for scene understanding in warehouse logistics, but they provide a limited number of images, along with corresponding 2D annotations. Furthermore, there are only a handful of general-purpose synthetic data simulators for generating photo-realistic images, like NVIDIA Isaac \cite{nvidia_developer_2020}, which provide warehouse scenes. However, there is no straightforward way to modify them to generate annotations needed for the task at hand. \looseness=-1
\\ \textbf{Domain Randomization:} 
We integrate domain randomization techniques \cite{tremblay2018training} in our dataset generation pipeline, as described in Sec. \ref{sec:racklay_dataset}. 


%% file: sections/method.tex


\section{Method}
\label{sec:method}
\subsection{Problem Formulation}
\label{sec:method:problemform}
Given a sequence of RGB images $\mathcal{I}_{1}$, $\mathcal{I}_{2}$,..., $\mathcal{I}_{n}$ of racks in warehouses in perspective view, we aim to predict the top-view (bird's eye view) and front-view layout for each rack present in each frame of the input video sequence.

\begin{figure}[!h]
    \center
    \includegraphics[width=\columnwidth]{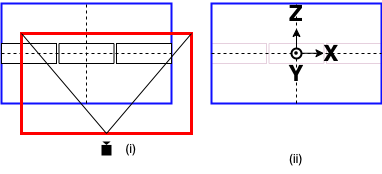}   \caption{\small{\textbf{Top-view representation} of the \textit{shelf-centric} (ii) layout for a given position of a shelf (i), and the reference coordinate frames for the same.}}
    \label{fig:topview}
    \vspace{-4mm}
\end{figure}

We consider \textbf{R} to be a rectangular area in a top-down orthographic view of the scene. The camera is placed at the mid-point of the lower side of the rectangle, directly facing the racks such that the image plane is orthogonal to the ground plane. (Fig. \ref{fig:topview}). Concretely, we want our network to generate top-view and front-view layouts for all the racks visible in each frame $\mathcal{I}_{t}$, within a region of interest $\Omega$. Our network predicts shelf-centric layouts where we map $\Omega$ to a rectangular area. In \textbf{shelf-centric} layout representation, we consider $\Omega$ to be a rectangular area, positioned such that its center coincides with the center of the shelves spanning across all the racks visible in the image, as shown in Fig. \ref{fig:topview}. This layout is hence with respect to the rack and is view-point agnostic. 

\begin{figure}[!h]
    \center
    \includegraphics[width=\columnwidth]{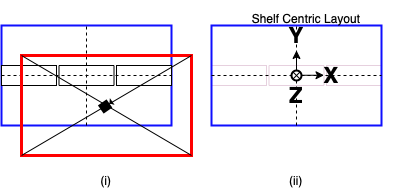}
    \caption{\textbf{Front-view representation} of the \textit{shelf-centric} (ii) layout for a given position of a shelf (i), and the reference coordinate frames for the same.}
    \label{fig:frontview}
    \vspace{-4mm}
\end{figure}

\begin{figure}[!h]
    \center
    \includegraphics[width=\columnwidth]{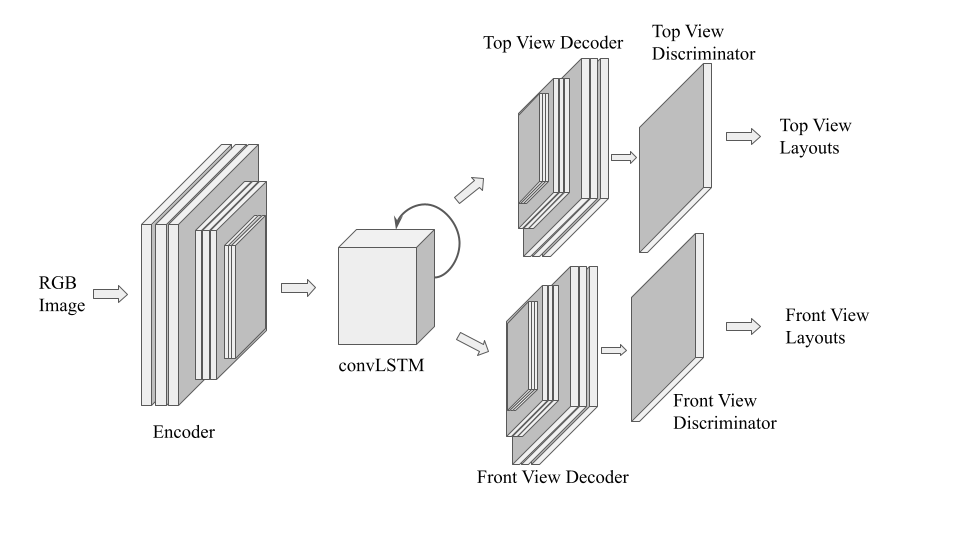}
    \caption{\textbf{Architecture:} The figure shows the architecture diagram for \textit{MVRackLay-disc}. It comprises a context encoder, a Convolutional LSTM for encoding temporal information, and multi-channel decoders and adversarial discriminators. (Refer to Sec. \ref{sec:method:arch}).}
    \label{fig:architecture}
    \vspace{-4mm}
\end{figure}

Our model predicts \textbf{top-view} and \textbf{front-view} layout representations for each frame in the sequence. Top-view layouts predict the bird's eye view occupancy of each shelf on the rack. Each pixel in the layout can either be classified as \textit{occupied}, \textit{unoccupied}, or \textit{background}. A pixel is said to be \textit{occupied} when it is a part of the object on the shelf, \textit{unoccupied} when it represents the empty space on the rack, and \textit{background} when it denotes the region which is not occupied by the shelf. 

Consider a right-handed coordinate frame where the X axis points to the right, the Y axis points downwards, and the Z axis points into the plane. For top-view layouts, the coordinate frame is positioned at the center of the shelves spanning across all racks for layouts. Hence, the top-view layouts are parallel to the X-Z plane, as in fig. \ref{fig:topview} (and the ground plane) at corresponding shelf heights. In the case of front-view layouts, the center of the coordinate frame is positioned at the center of shelves in X and Y directions. Front view layouts are, therefore, orthogonal to the ground plane, as in fig. \ref{fig:frontview}.



As an additional task, we demonstrate \textbf{Multi-view Stitching} - combining the representation from the top-view and front-view layouts to obtain a 3D reconstruction of all racks in the warehouse in the global shelf frame (Fig. \ref{fig:multi}). This can be further used for 3D spatial reasoning tasks. We infer the X and Z coordinates from the top-view and the Y coordinate from the front-view. We further explore these applications in Sec. \ref{sec:experiments:applications}.

\subsection{MVRackLay Architecture}
\label{sec:method:arch}

We build a double-decoder MVRackLay architecture (Fig. \ref{fig:architecture}) which takes as input a sequence of RGB images $\mathcal{I}_{1}$, $\mathcal{I}_{2}$,..., $\mathcal{I}_{n}$ and predicts the top-view and front-view layouts. The components of the model are described in detail below.

     \noindent \textbf{1)} A \textbf{context encoder} which uses a ResNet-18 backbone pre-trained on ImageNet \cite{deng2009imagenet} to retrieve relevant 3D scene and semantic components from the monocular input $\mathcal{I}_{t}$. This feature extractor learns low-level features $\mathcal{C}_{1}$, $\mathcal{C}_{2}$,..., $\mathcal{C}_{n}$ that help reason about the occupancy of the scene points.\looseness=-1
    
    \noindent \textbf{2)} A stacked \textbf{Convolutional LSTM} submodule uses the encoder extracted features $\mathcal{C}_{1}$, $\mathcal{C}_{2}$,..., $\mathcal{C}_{t}$ and in turn encodes a temporal representation to capture motion across input frames. We use this Spatio-temporal prediction to estimate consistent layouts by consolidating the information from the past frames to predict the current frame. The number of previous frames used in this prediction is a hyperparameter, the value of which is varied in our ablation studies (Refer to Sec. \ref{sec:experiments:ablation}). The output of this block is an encoded representation that better reasons the scene points as \textit{occupied}, \textit{unoccupied}, and \textit{background}.
    
    \noindent \textbf{3)} A \textbf{top-view decoder} and \textbf{front-view decoder} that generates layouts respectively for top-view and front-view from the temporal representation learned by the Convolutional LSTM submodule. It consists of downsampling layers to output an $\mathcal{R}$ × $\mathcal{D}$ × $\mathcal{D}$ grid which represents the layout, where $\mathcal{R}$ is the number of output channels and $\mathcal{D}$ is the resolution of the output layouts. 
    
     \noindent \textbf{4)} Identical \textbf{discriminators} following top-view and front-view decoders, respectively, are adversarial regularizers that rectify the layouts further by homogenizing their distributions to be similar to the true distribution of plausible layouts. The layout predicted by the decoder is the input to this submodule which outputs the final refined predictions. 
     \looseness=-1
\subsection{Loss function}
We describe here the loss function of our \textit{MVRackLay} architecture for top-view layout estimation. We use stochastic gradient descent to optimize over the network parameters $\phi$, $\psi$, $\theta$ of the context encoder, convolutional LSTM, and the decoder.
\vspace{-1mm}

\[{{L}_{sup}(\widehat{\mathcal{T}};\phi, \psi) = \sum_{j=1}^{N}\sum_{i=1}^{\mathcal{R}} f\left( \widehat{\mathcal{T}}_{i}^{j}, \mathcal{T}_{i}^{j} \right)}\]

\[{{L}_{adv}(\widehat{\mathcal{T}};\phi, \psi, \theta) = \E_{\theta\sim p_{fake}}[(\widehat{\mathcal{T}}(\theta)-1)^{2}] }\]

 \[{{L}_{short}(\widehat{\mathcal{T}};\phi, \psi) = \sum_{j=1}^{N}\sum_{i=1}^{{seqlen-1}} f\left( \widehat{\mathcal{T}}_{i}^{j}, \widehat{\mathcal{T}}_{i}^{j+1} \right)}\]
 
\[{{L}_{long}(\widehat{\mathcal{T}};\phi, \psi) = \sum_{j=1}^{N}\sum_{i=1}^{seqlen - 1}\sum_{k=j+2}^{seqlen} f\left( \widehat{\mathcal{T}}_{i}^{j}, \widehat{\mathcal{T}}_{i}^{k} \right)}\]

\[{{L}_{discr}(\widehat{\mathcal{T}};\theta) = \E_{\theta\sim p_{true}}[(\widehat{\mathcal{T}}(\theta)-1)^{2}] \\ + \E_{\theta\sim p_{fake}}[ (\widehat{\mathcal{T}}(\theta))^{2}]}\]

\[
 {{L}_{total} = {L}_{sup} + {L}_{short} + {L}_{long} + {L}_{adv} + {L}_{discr}}\]

\noindent Here $\widehat{\mathcal{T}}$ is the predicted top-view layout of each shelf, $\mathcal{T}$ is the ground truth top-view layout of each shelf, $\mathcal{R}$ is the maximum number of shelves considered, and $N$ is the mini-batch size.

 \textbf{${L}_{sup}$} is the per-pixel cross-entropy loss which penalizes variation of the predicted output labels ($\widehat{\mathcal{T}}$) from corresponding ground-truth values ($\mathcal{T}$). The adversarial loss ${L}_{adv}$ enables the distribution of layout estimates from the top-view decoder ($p_{fake}$) to be similar to the actual data distribution ($p_{true}$). ${L}_{discr}$ enforces the discriminator to accurately classify the network-generated top-view layouts sampled from the true data distribution \cite{GAN}. ${L}_{short}$ is the short-range consistency loss, and ${L}_{long}$ is the long-range consistency loss. Finally, we minimize the total loss over the network parameters $\phi$, $\psi$, $\theta$ and use it to back-propagate gradients through the network. Equivalent expressions are defined for front-view layout prediction as well.

%% file: sections/dataset.tex
\begin{table}[!ht]
\begin{center}
\begin{adjustbox}{width=\columnwidth}
\begin{tabular}{|c|c|c|c|c|c|c|c|c|}
\hline
& \multicolumn{4}{c|}{\textbf{Top View}} 
& \multicolumn{4}{c|}{\textbf{Front View}} \\ 
\cline{2-9}
& \multicolumn{2}{c|}{\textbf{Rack}} 
&   \multicolumn{2}{c|}{\textbf{Box}}    
& \multicolumn{2}{c|}{\textbf{Rack}}   
& \multicolumn{2}{c|}{\textbf{Box}}  \\ 
\hline
\textbf{Method}       & \textbf{mIoU}      &     \textbf{mAP}           &  \textbf{mIoU}       & \textbf{mAP}         &   \textbf{mIoU}    &   \textbf{mAP}   &      \textbf{mIoU}   &   \textbf{mAP}   \\
\cline{1-9}
MVRackLay-Disc-4 (ours)  & $\textbf{96.44}$   &      $98.01$      & $\textbf{86.89}$     &  $\textbf{92.70}$    &   $\textbf{94.98}$  &   $\textbf{97.14}$   &   $\textbf{88.02}$  &  $\textbf{93.49}$     \\ 
MVRackLay-Disc-8 (ours) & $95.84$   &      $\textbf{98.12}$      & $85.98$     &  $88.35$    &  ${93.78}$  &  ${96.98}$   &   $86.49$   &  $88.76$         \\

MVRackLay-4 (ours)   & ${95.94}$   &      $97.94$      & ${86.66}$     &  ${89.31}$    &   $93.73$  &   $96.58$   &   ${86.72}$  &  ${89.16}$     \\ 

RackLay-D-disc  & ${93.44}$   &      ${94.98}$      & ${82.80}$     &  ${85.47}$      &   ${91.75}$  &  ${93.10}$   &   $83.28$  &  $85.49$     \\ 

\cline{1-9} 
PseudoLidar-PointRCNN\cite{shi2019pointrcnn}   & $73.28$   &      $77.40$      &  $55.77$     &  $81.26$    &   $-$   & $-$   &     $63.05$   &  $89.45$     \\

MaskRCNN-GTdepth\cite{he2017mask}   &   $-$   &     $-$      &  $35.57 $     &  $47.44$    &   $-$   & $-$   &     $76.48$   &  $82.48$     \\
\cline{1-9}

\hline
\end{tabular}
\end{adjustbox} 
\end{center}
\caption{\textbf{Quantitative results}:
We benchmark the 3 different versions of our network - \textit{MVRackLay-Disc-4}, \textit{MVRackLay-Disc-8} and \textit{MVRackLay-4}, along with three baselines- \textit{RackLay-D-disc}\cite{racklay} \textit{PseudoLidar-PointRCNN}\cite{wang2019pseudo,shi2019pointrcnn} and \textit{MaskRCNN-GTdepth}\cite{he2017mask} (as described in Sec. \ref{sec:experiments:eval}).}
\label{table:quantitative:main}
\vspace{-3mm}
\end{table}

\section{Dataset Generation Pipeline}
\label{sec:dataset}

In this section, we introduce \textit{WareSynth}\footnote{For more information on \textit{WareSynth} and code: \url{https://pranjali-pathre.github.io/MVRackLay/}}, a robust pipeline for synthetic data generation, inspired by the more simplistic pipeline in \cite{racklay}. \textit{WareSynth} is an end-to-end pipeline that can be used to generate 3D warehouse scenes, capture the relevant data and output the annotations. It is developed in the 3D graphics framework Unity \cite{unity} in order to generate more realistic images compared to the original pipeline created in Blender \cite{blender}. By using the \textit{WareSynth} pipeline on an NVIDIA RTX 2080Ti, we are able to generate 500 images per minute. Our pipeline is characterized by the following:
\begin{enumerate}
    \item Highly customizable as per user requirements.
    \item Ability to export to popular annotation formats such as COCO, YOLO, Pix3D, KITTI and BOP.
    \item Completely free and open source.
    \item Extensive variation and domain randomization.
\end{enumerate}

Although only a handful of  warehouse datasets/simulators are currently available such as LOCO and NVIDIA Isaac (discussed in Sec.  \ref{sec:relatedwork:dataset}), to the best of our knowledge, there is no such pipeline that satisfies all four of the above criteria. \textit{WareSynth} allows users to customize the number of racks and boxes in the scene and also the box and rack models used. The ability to export to not only our annotation format but also popular annotation formats is important for additional benchmarks against new approaches being developed in these settings. Section \ref{sec:racklay_dataset} enumerates the randomizations enabled by \textit{WareSynth}. Our data generation and capture methods are efficient, flexible, and easily customizable based on user requirements. Hence, \textit{WareSynth} can prove very useful for large-scale data annotation generation.\looseness=-1

%% file: sections/experiments.tex
\section{Experiments and Analysis}
\label{sec:experiments}

\subsection{MVRackLay Dataset}
\label{sec:racklay_dataset}
For training and testing our network, we generated a diverse dataset with 20k images spanning 400 sequences with 50 images per sequence, using \textit{WareSynth}\footnote{Download RackLay dataset: \url{https://tinyurl.com/yxmu5t64}}. 
It is split into 360/20/20 for train/test/validation. All the results discussed are on the test set of this dataset. The dataset is highly varied and spans multiple warehouse scenes. The variety demonstrates the generality of \textit{MVRackLay} and is useful to evaluate the performance of our model in varied warehouse settings. The video sequences captured using \textit{WareSynth} resemble the data captured by a manual camera movement or a mechanized system performing the task in an actual warehouse. 

Various scene elements were diversified during data generation to impart assortment in generated scenes so that the synthetically developed warehouses mimic their real-world counterparts. We describe these randomizations below.

\label{sec:racklay_dataset:domain_randomization}
\noindent \textbf{Domain Randomization:} We show the randomizations we introduce using 3 randomly selected images from our dataset through Fig. \ref{fig:final_results} (referred throughout this section): 

\begin{itemize}
    \item Boxes have random sizes, textures, rotation about the vertical axis, colors, and reflective properties.
    \item Box placement varies from dense to moderate to sparse.
    \item Color and texture of racks are randomized.
    \item Height to which boxes are stacked vertically is randomized.
    \item Background is either a wall or a busy warehouse.
    \item Color and textures of floors and walls are randomized.
    \item The camera's position concerning the rack varies within a range to capture different numbers of shelves in the sequence. For our dataset, we set $\mathcal{R}$=3.
    \item Camera is moved such that different number of racks are visible across frames.
\end{itemize}

\textit{We find that this large diversity in the dataset has enabled the network to not overfit on the domain of the synthetic data, but rather learn features that emulate real-world scenes}.

\subsection{Evaluated Methods and Metrics}
\label{sec:experiments:eval}

We compare the following variants of MVRackLay:

\begin{itemize}
    \item \textit{MVRackLay-Disc-4}: Double decoder architecture with discriminators for both front-view \textit{and} top-view, with a sequence length of 4 used in the ConvLSTM module.
    \item \textit{MVRackLay-Disc-8}: Double decoder architecture with discriminators for both front-view \textit{and} top-view, with a sequence length of 8 used in the ConvLSTM module.
    \item \textit{MVRackLay-4}: Double decoder architecture for both front-view \textit{and} top-view without discriminators, with a sequence length of 4 used in the ConvLSTM module.
\end{itemize}
We report Mean Intersection-Over-Union (mIoU) and Mean Average-Precision (mAP) scores in this task as the previously proposed methods also evaluate the model on the same criterion.   
\subsection{Results \protect\footnote{More results: \url{https://pranjali-pathre.github.io/MVRackLay/}}}
\label{sec:experiments:results}

We first trained \textit{MVRackLay-4} for top-view and front-view. Having achieved superior results compared to baselines, we further trained \textit{MVRackLay-Disc-4} for both front-view and top-view to capture the distribution of our layouts, which led to the best results. We observed performance gains as discussed in Sec. \ref{sec:experiments:ablation}. We further trained \textit{MVRackLay-Disc-8} to capture the performance of the model for higher sequence length (discussed in Sec. \ref{sec:experiments:ablation}). Overall, \textit{MVRackLay-Disc-4} showed the best overall performance (refer Table \ref{table:quantitative:main}).

\begin{figure}[!t]
\centering
\includegraphics[width=1.0\columnwidth]{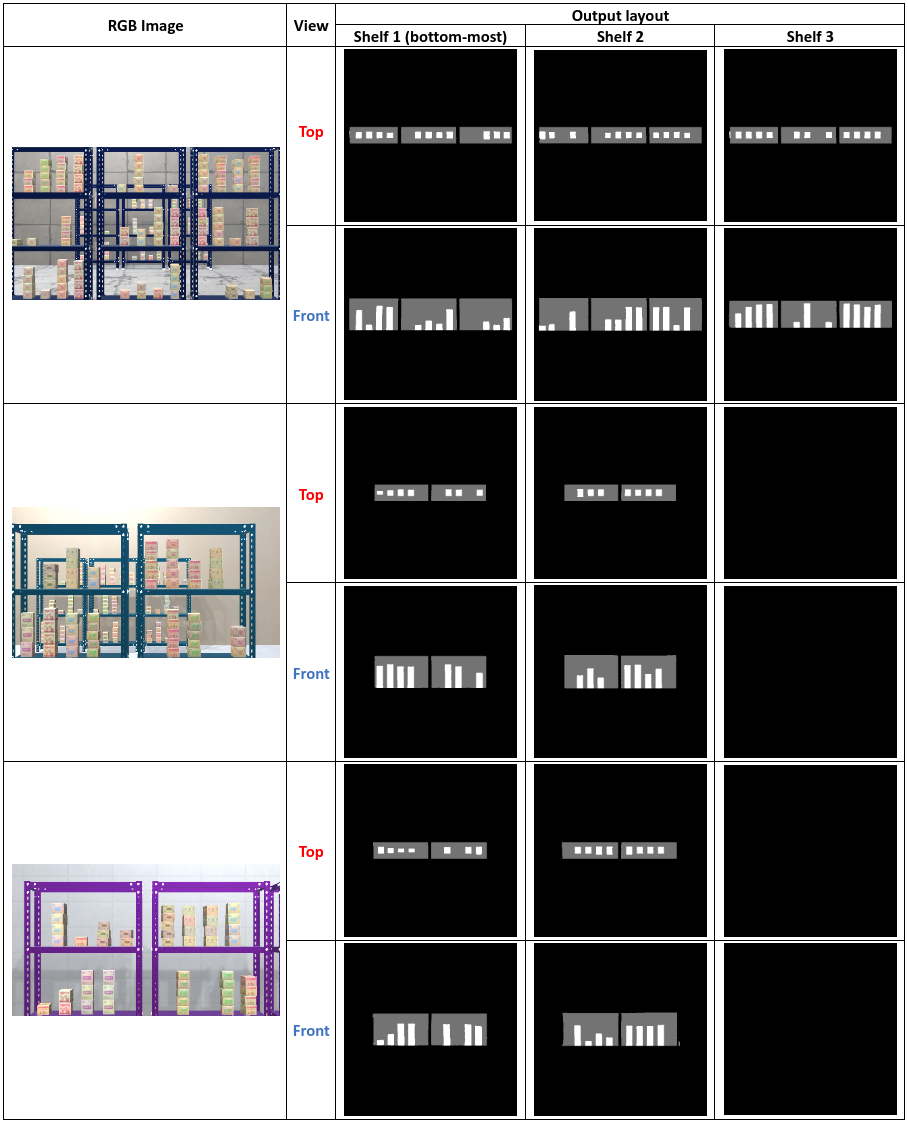}\caption{\textbf{\textit{MVRackLay-Disc-4} Results}: Here, we present the results of our network tested on domain randomized data. The bottom-most shelf layout is shown in the left-most column, followed by the middle and top shelf (if visible). Observe the diversity of warehouse scenes captured (detailed in \ref{sec:racklay_dataset:domain_randomization}) and the top-view and front-view layouts predicted for the same.}
\label{fig:final_results}
\vspace{-7mm}
\end{figure}

Fig. \ref{fig:final_results} summarizes the results of \textit{MVRackLay-disc-4} tested on domain randomized data. Our best network is able to predict all the racks present in the image with clean boundaries separating them and precisely estimate the space between two racks and two objects on the shelf. \textit{MVRackLay} can rightly predict the layouts for the racks in the foreground, even in the presence of background clutter in the image. It can also reason for the narrow spaces in densely packed shelves. 

The results show that \textit{MVRackLay} can significantly benefit downstream warehouse tasks. Sec. \ref{sec:experiments:applications} demonstrates one such task of 3D warehouse reconstruction using a multiview layout predicted by the network. The generality and superiority of the results show that our model can easily be transferred to warehouses with diverse arrangements of racks and objects on shelves.

\subsection{Comparison with baselines}
\label{sec:experiments:baselines}
\begin{figure}[!t]
\centering
\includegraphics[width=1.0\columnwidth]{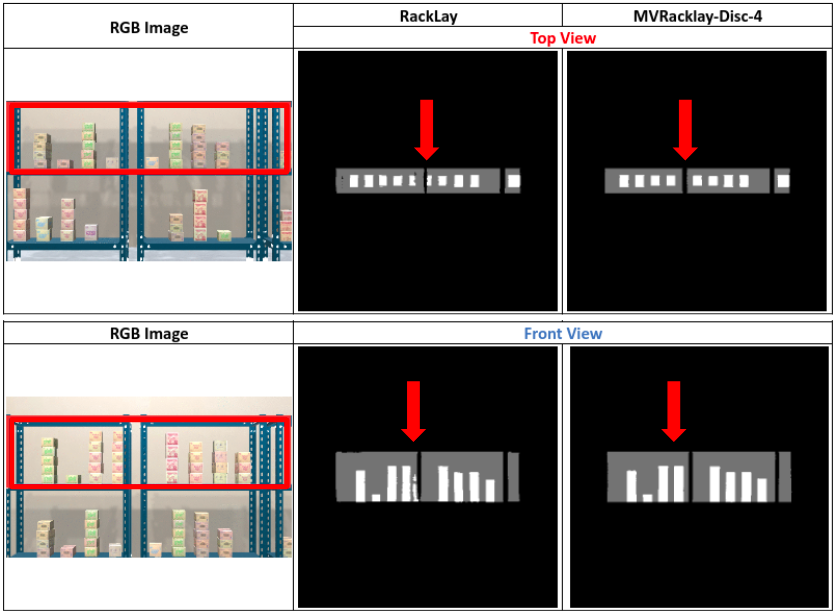}\caption{\textbf{\textit{RackLay} vs. \textit{MVRackLay-Disc-4}}: Above, we compare qualitatively the results of \textit{RackLay} and our \textit{MVRackLay-Disc-4}. The shelf in focus is highlighted with a red border. Observe that our model removes the false positive in row 1, removes noise in row 2, and increases the sharpness of both box boundaries (both rows) and shelf edges (row 2).}.
\label{fig:mvracklay_vs_racklay}
\vspace{-7mm}
\end{figure}

\noindent \textbf{RackLay:} RackLay \cite{racklay} was proposed to solve the problem of layout prediction for the dominant rack present in the input RGB image. We trained it for the task of layout prediction for all the racks present over the video sequence. From Table \ref{table:quantitative:main} it is clear that \textit{MVRackLay-Disc-4} performs significantly better than \textit{RackLay-D-disc} quantitatively. Comparing qualitatively, Fig. \ref{fig:mvracklay_vs_racklay} summarizes the improvements of our network over \textit{RackLay}. \textit{RackLay} often fails to demarcate an exact boundary between two racks present in an image (row 1). In row 2, observe that \textit{RackLay} is unable to predict a sharp boundary of both box and shelf as in the corresponding output of \textit{MVRackLay-Disc-4}. \textit{RackLay} often incorrectly predicts the presence of the box on racks (rows 1, 2) or suffers from noisy predictions of boxes in regions with no objects as well as of the shelf (row 2). Our model not only extends \textit{RackLay}'s functionality to image sequences but also improves upon its liabilities. We predict sharper and more accurate shelf and object boundaries and reduce false box predictions.

\noindent \textbf{PseudoLidar-PointRCNN:}
PointRCNN \cite{shi2019pointrcnn} is a  3D object detector that uses the raw point cloud as input. Hence we use the PseudoLidar\cite{wang2019pseudo} information to detect 3D objects using PointRCNN. As this method considers a single dominant plane and is employed for birds-eye view prediction, we report metrics for the bottommost shelf (refer to Table \ref{table:quantitative:main}) and the top-view prediction only. Accounting for the single-dominant layer assumption, it is clear from Table \ref{table:quantitative:main} that our model performs better as it performs multilayer, front-view, and top-view layout predictions that can also reason about the inter-shelf distance.

\noindent \textbf{Mask R-CNN:} 
We select Mask R-CNN\cite{he2017mask} as one of our baselines to test the instance segmentation method for multi-layer layout prediction in the warehouse setting for the sequential data. We subsequently integrate the Mask R-CNN segmented instances with the depth maps and project on a horizontal plane to segment the boxes shelf-wise, using the fact that a set of boxes on a particular shelf will be situated on the same plane located at some elevation from the ground plane. From the experiment, it is observed that Mask R-CNN fails to detect the precise boundary of the rack due to its thin structures. If multiple racks are present in the image, Mask R-CNN also fails to mark a clear distinction between them. The results summarized in Table \ref{table:quantitative:main} prove that our model performs better quantitatively too. Since Mask R-CNN can only claim regarding the points visible in the image, it is evident that our model accomplishes better results with amodal estimation to reason beyond the visual elements that structurally characterize racks and objects. 



\subsection{Ablation studies}
\label{sec:experiments:ablation}
To thoroughly comprehend the underlying effect of different components, we perform ablation studies over the model's architecture and examine its effect on performance. 

\begin{figure}[!t]
\centering
\includegraphics[width=1.0\columnwidth]{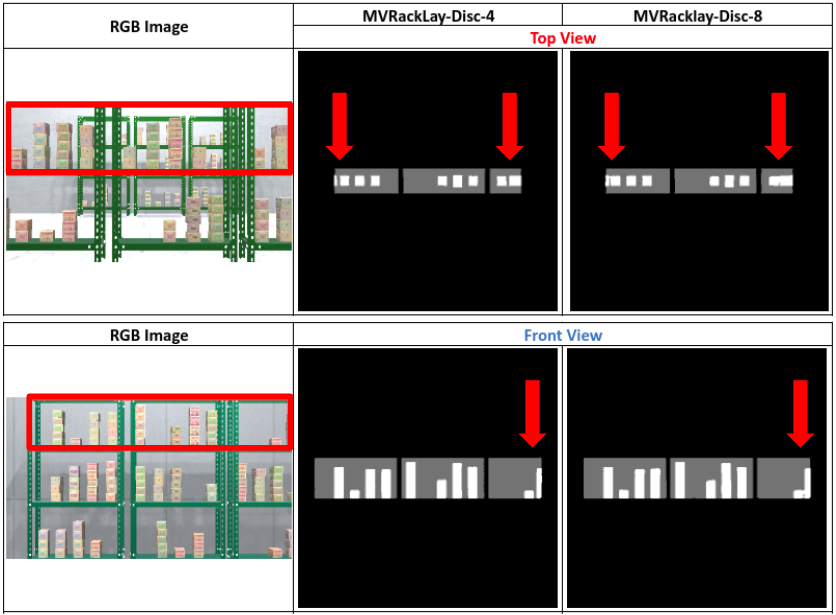}\caption{\textbf{\textit{MVRackLay-Disc-4} vs. \textit{MVRackLay-Disc-8}}: The shelf in focus is highlighted with a red border. Better demarcations between adjoining boxes and less joining of abreast layouts are observed in the output of \textit{MVRackLay-Disc-4} compared to its counterpart.}
\label{fig:mvracklay_4vs8}
\vspace{-7mm}
\end{figure}

\begin{figure}[!t]
\centering
\includegraphics[width=1.0\columnwidth]{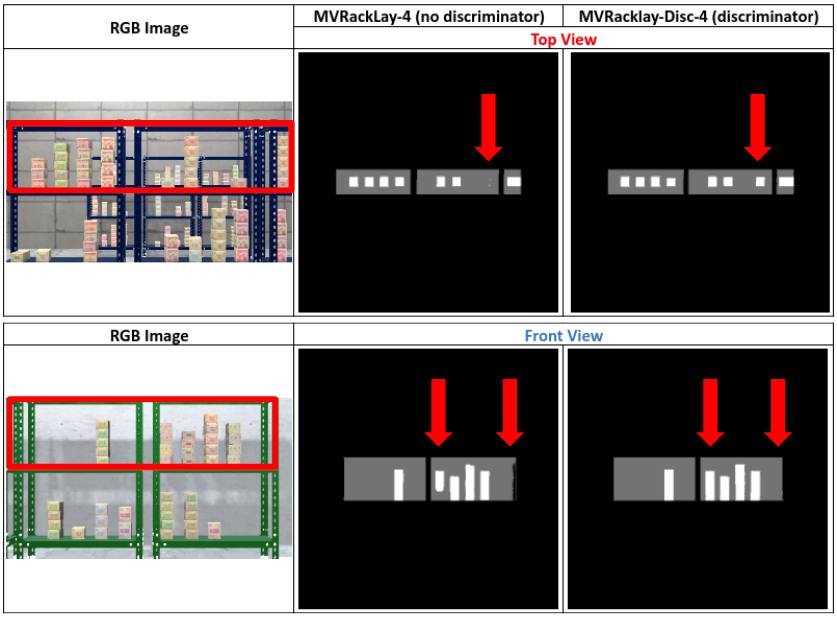}\caption{\textbf{\textit{MVRackLay-4} vs. \textit{MVRackLay-Disc-4}}: The shelf in focus is highlighted with a red border. Observe how using a discriminator fixes the false negative in row 1 and improves predicted box boundaries and shelf boundaries in row 2.}
\label{fig:mvracklay_disc_vs_no_disc}
\vspace{-3mm}
\end{figure}

\noindent \textbf{1) Convolutional LSTM sequence length:} We varied the time steps used in the stacked Convolutional LSTM submodule. We observed that \textit{MVRackLay-Disc-4} was able to converge faster and improve qualitatively over \textit{MVRackLay-Disc-8}. Although quantitatively  \textit{MVRackLay-Disc-4} and  \textit{MVRackLay-Disc-8} perform alike (refer to Table \ref{table:quantitative:main}), from fig. \ref{fig:mvracklay_4vs8} considerable qualitative improvements can be observed. \textit{MVRackLay-Disc-4} performs better in identifying precise object boundaries and distinguishing between the closely spaced objects on the rack. A lower sequence length enabled the model to compile only relevant details from past frames and avoid spurious noise. 

\noindent \textbf{2) Adversarial learning:}
In \textit{MVRackLay-Disc-4}, we add discriminators after decoders in \textit{MVRackLay-4} to capture the distribution of \textit{plausible} layouts. We observed a substantial improvement both quantitatively (refer to Table \ref{table:quantitative:main}) and qualitatively (Fig. \ref{fig:mvracklay_disc_vs_no_disc}). Layouts have become sharper, and most notably, using a discriminator diminished the stray pixels wrongly categorized as boxes. The actual distance between the boxes positioned near the end of the shelf is difficult to estimate as they are imaged obliquely. In such cases, \textit{MVRackLay-Disc-4} remarkably improved the prediction of the boxes and generated cleaner layouts.  

\subsection{Applications} 
\label{sec:experiments:applications}

\noindent \textbf{Multi View Stitching:} From the layout prediction of a particular frame $\mathcal{I}_{t}$, we first obtain the 2D bounding boxes of all shelves and boxes detected in the front-view and top-view layout. The detections from the top-view and front-view layouts are corresponded to identify the matching. Once we have a map, using the dimension information from front-view and top-view layout predictions, we generate the 3D bounding box for all the mapped racks and objects. Finally, we combine these representations of each shelf to get a 3D reconstruction ${f}_{t}$ of all the racks in the frame. This process is repeated for all the frames in the sequence.

Given two consecutive 3D reconstructions ${f}_{t}$ and ${f}_{t+1}$, we initially find all the corresponding matching boxes. We then calculate the shift between them; using this shift, we discern the direction of the motion. Finally, we consider the last box in frame ${f}_{t}$ in the direction of motion and check its corresponding box in frame ${f}_{t+1}$. If the size of the box in ${f}_{t+1}$ is larger, we increase the size of the shelf and boxes in ${f}_{t}$ accordingly. If there is an addition of a new box or shelf in ${f}_{t+1}$, the same is included in the ${f}_{t}$ reconstruction. Eventually, we obtain the merged layouts of ${f}_{t}$ and ${f}_{t+1}$ in ${f}_{t}$ frame.

\begin{figure}[!h]
\centering
\includegraphics[width=\columnwidth]{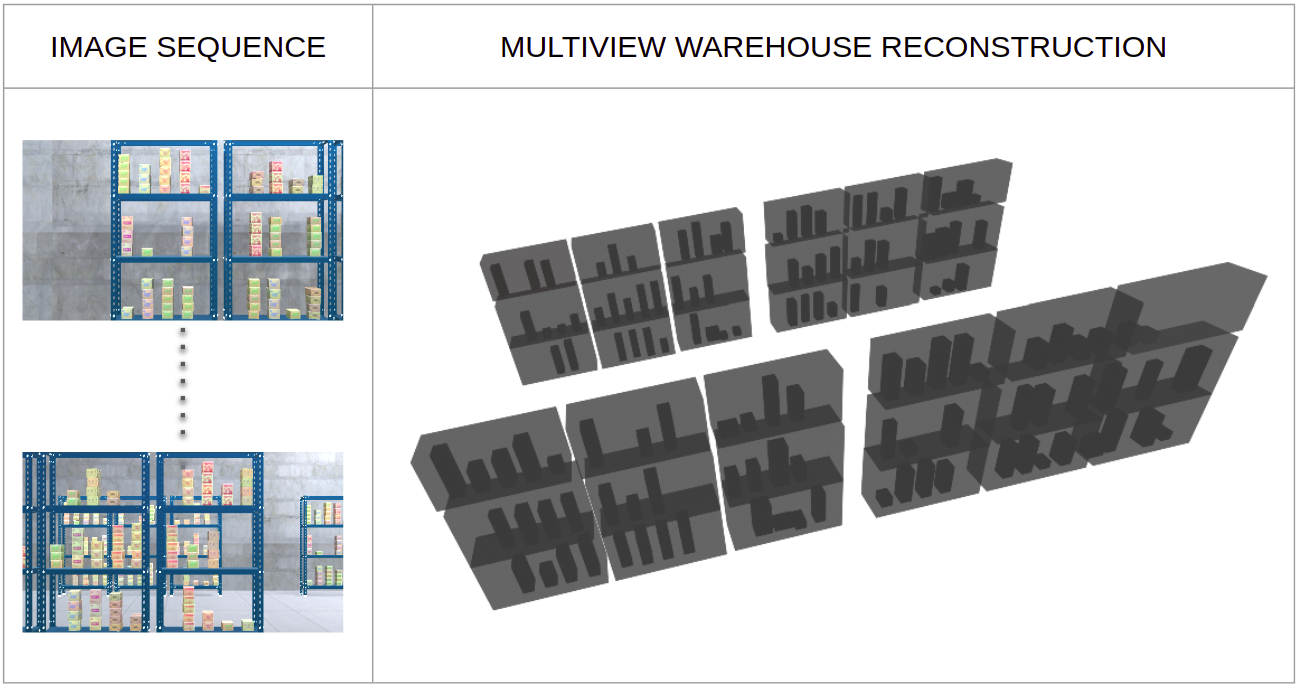}
\caption{\textbf{Multi View Reconstruction} of the entire warehouse using four sequences covering 280 frames (70 frames each), using the layouts predicted by \textit{MVRackLay-Disc-4}.}
\label{fig:multi}
\vspace{-4mm}
\end{figure}

If ${F}_{t}$ denotes the merged 3D representation from ${f}_{1}$ to ${f}_{t}$, ${f}_{t+1}$ is merged into ${F}_{t}$ using the method described above. Fig. \ref{fig:multi} shows the 3D reconstruction of a single warehouse with 4 racks, obtained from the multi-view stitching of predicted layouts of 4 sequences with 70 frames per sequence.

%% file: sections/conclusion.tex
\section{Conclusion}
In this paper, we present \textit{MVRackLay} to perform multi-view \underline{and} multi-layered layout estimation for all racks partly or fully visible in each frame of the input monocular image sequence. Distinct from existing methods, it utilizes temporal information across the frames of a image sequence to enhance the quality of layouts. We also present a pipeline to 3D reconstruct the entire warehouse from the predicted shelf-centric layouts. Further, we introduce a versatile synthetic data generation pipeline, \textit{WareSynth}, that is capable of producing domain randomized data which can emulate a wide variety of warehouse scenes. \textit{MVRackLay}'s versatility is demonstrated by the experimental results over diverse warehouse scenes, and is vastly superior to previous baselines adapted for the same task.